\documentclass[sigconf,nonacm]{acmart}

\setcopyright{none}
\renewcommand\footnotetextcopyrightpermission[1]{}
\usepackage{booktabs}
\usepackage{graphicx}
\usepackage{microtype}
\usepackage{url}
\graphicspath{{figures/}}
\title{Temporal Heterogeneous Graph Transformer for Credit Card Fraud Detection}

\author{Qinwen Yan}
\email{florayan@g.ucla.edu}
\affiliation{%
  \institution{University of California, Los Angeles}
  \city{Los Angeles}
  \state{California}
  \country{USA}
}

\begin{document}

\begin{abstract}
Credit card fraud detection typically relies on tabular features, while repeated attributes can also provide useful relational signals. This paper proposes THGT-FD, a Temporal Heterogeneous Graph Transformer for Fraud Detection. Each transaction is represented using one transaction token and six types of relation tokens and incorporates Time2Vec encoding into the transaction representation. A Transformer learns the interactions among these tokens within each individual transaction and then outputs a fraud probability. Experiments were conducted on 150,000 transactions sampled from the IEEE-CIS Fraud Detection dataset and chronologically partitioned according to \texttt{TransactionDT}. On the test set, THGT-FD achieved an AUC-ROC of 0.8536, an average precision of 0.4164, and a Recall@5\% of 0.4708. The class-weighted histogram-based gradient-boosting baseline achieved an AUC-ROC of 0.8722. The results indicate that relation tokens provide useful information for fraud-risk ranking, although the current model does not yet incorporate entity-level historical aggregation.
\end{abstract}

\keywords{fraud detection, heterogeneous graph, Transformer, Time2Vec, class imbalance}

\maketitle

\section{Introduction}
The continued expansion of online payments has placed increasing pressure on financial institutions to identify fraudulent transactions. Fraudulent transactions usually account for a small proportion of all transactions and may closely resemble legitimate ones. Fraud detection models that rely solely on individual transaction records, such as transaction amounts, often struggle to detect risk in a timely manner. Some risk patterns will not become apparent until repeated attributes associated with transactions are considered. Therefore, effectively leveraging relational signals in high-dimensional transaction data has become an important research problem in credit card fraud detection~\cite{bolton2002review,dalpozzolo2018realistic,jurgovsky2018sequence}.

The IEEE-CIS Fraud Detection dataset provides a challenging experimental setting for solving this problem~\cite{ieeecis}; it contains many high-dimensional anonymized variables and substantial missing data. Recent studies on this benchmark have consequently examined high-cardinality categorical encoders and generative priors for noisy, imbalanced transactions~\cite{han2026interpretableencoders,xu2026generative}. Gradient-boosting models can effectively capture nonlinear interactions, making them strong baselines for tabular data~\cite{friedman2001gbm,chen2016xgboost,pedregosa2011sklearn}. However, these models mostly generate predictions from individual transaction records. Although repeated attributes can be encoded as categorical features, their underlying relational meaning is only indirectly represented. More explicit relational information generally requires manually engineered aggregation features~\cite{whitrow2009aggregation}.

Graph neural networks provide an alternative approach to modeling transaction relationships. GCN and GraphSAGE learn node representations by aggregating information from neighboring nodes~\cite{kipf2017gcn,hamilton2017graphsage}. Network-based fraud detection, R-GCN, and heterogeneous graph transformers further account for relational structure or differences among relation types~\cite{vanvlasselaer2015apate,schlichtkrull2018rgcn,hu2020hgt}. Nevertheless, in anonymized transaction data, repeated values do not necessarily correspond to verified real-world entities. Connections with ambiguous meanings may reduce representation quality, while constructing a complete transaction graph can substantially increase computational costs. Temporal modeling introduces an additional challenge, as fraud risk may evolve with transaction order and may also be influenced by the pace of transaction activity; recent financial-risk studies have addressed related dependencies using improved sequence models and joint Transformer--graph architectures~\cite{xu2025mamba,liang2026spatiotemporal}.

Here, we propose THGT-FD, a Temporal Heterogeneous Graph-Token Transformer for credit card fraud detection. Each transaction is represented by one transaction token and six relation tokens derived from anonymized fields. Time2Vec is used to encode the relative temporal position of each transaction~\cite{kazemi2019time2vec}, while a Transformer encoder learns interactions between the transaction token and relation tokens~\cite{vaswani2017attention}. Using the IEEE-CIS dataset as a case study, model performance is evaluated with fraud-risk ranking metrics, with a class-weighted histogram-based gradient-boosting model serving as the baseline. Overall, this study presents a compact approach to modeling relational information in anonymized transaction data, thereby improving the framework's ability to capture evolving behavioral patterns and support more robust fraud detection.

\section{Method}
\subsection{Data Processing and Relation Tokens}
The transaction table contains the target label, \texttt{TransactionDT}, \texttt{TransactionAmt}, product codes, card attributes, and address attributes. The identity table contains device and browser information. The merged data are temporally divided into training, validation, and test sets. All preprocessing parameters are estimated from the training set.

Features whose missing rates exceed a predefined threshold are removed. Numerical features are imputed with training-set medians and standardized using training-set means and standard deviations. Categorical features not used to construct relation tokens are frequency-encoded. For a categorical value $c$, the encoding is defined as
\begin{equation}
f(c)=\log\left[1+N_{\mathrm{train}}(c)\right],
\label{eq:frequency}
\end{equation}
where $N_{\mathrm{train}}(c)$ denotes the frequency of $c$ in the training set. Relative day and hour features are also derived from \texttt{TransactionDT}.

Six relation types are constructed. The card relation is jointly defined by \texttt{card1}--\texttt{card6}, and the address relation by \texttt{addr1} and \texttt{addr2}. The product relation is defined by \texttt{ProductCD}. Purchaser and recipient email domains form two separate relations. The device relation is jointly defined by \texttt{DeviceType} and \texttt{DeviceInfo}.

Each relation value is mapped to an integer identifier and transformed using a relation-specific embedding table. For relation type $k$ of transaction $i$, the relation token is computed as
\begin{equation}
\mathbf{z}_i^k=\mathbf{E}_k(r_i^k)+\mathbf{e}_k^{\mathrm{type}},
\label{eq:relation-token}
\end{equation}
where $r_i^k$ is the relation identifier, $\mathbf{E}_k$ is the embedding table for relation type $k$, and $\mathbf{e}_k^{\mathrm{type}}$ is the corresponding type embedding.

\subsection{Model Architecture and Training}
The transaction token is constructed from the processed transaction features and temporal representation. The normalized timestamp $t_i$ is encoded using Time2Vec with one linear component and $m$ periodic components:
\begin{equation}
\operatorname{T2V}(t_i)=\left[w_0t_i+b_0,\,\sin(w_1t_i+b_1),\ldots,\sin(w_mt_i+b_m)\right].
\label{eq:time2vec}
\end{equation}
The transaction features $\mathbf{x}_i$ are concatenated with the temporal representation and projected into the hidden space:
\begin{equation}
\mathbf{z}_i^{\mathrm{tr}}=\mathbf{W}_{\mathrm{tr}}[\mathbf{x}_i\Vert\operatorname{T2V}(t_i)]+\mathbf{b}_{\mathrm{tr}}.
\label{eq:transaction-token}
\end{equation}

The input sequence $\mathbf{Z}_i$ consists of one transaction token and six relation tokens in a fixed order. It is processed by a two-layer Transformer encoder with four attention heads per layer. Let $\mathbf{H}_i$ denote the encoder output. The representation at the first position, $\mathbf{H}_i^0$, is passed through a multilayer perceptron to estimate the fraud probability:
\begin{equation}
\mathbf{H}_i=\operatorname{Transformer}(\mathbf{Z}_i),\qquad
p_i=\sigma\!\left[\operatorname{MLP}(\mathbf{H}_i^0)\right].
\label{eq:prediction}
\end{equation}

\begin{figure*}[t]
  \centering
  \includegraphics[width=0.98\textwidth]{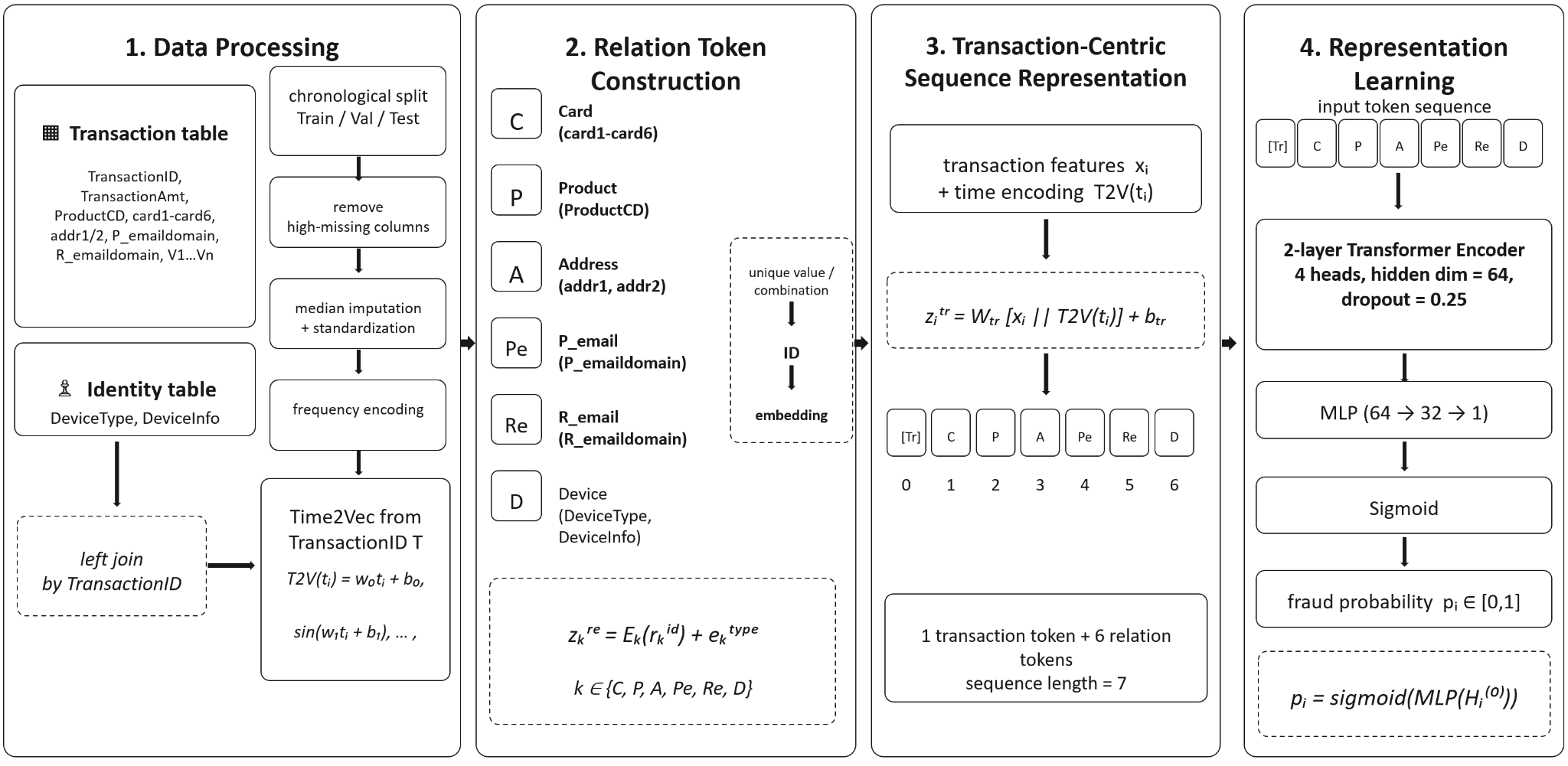}
  \Description{A four-stage THGT-FD workflow covering data processing, relation-token construction, transaction-centric sequence representation, and Transformer-based representation learning.}
  \caption{Transaction-centered heterogeneous graph representation and prediction pipeline of THGT-FD.}
  \label{fig:method}
\end{figure*}

The model is trained using AdamW~\cite{loshchilov2019adamw} and class-weighted focal loss~\cite{lin2017focal,he2009imbalance}. Let $p_t$ denote the predicted probability of the ground-truth class. The loss for each sample is defined as
\begin{equation}
\mathcal{L}=-w(y)(1-p_t)^{\gamma}\log(p_t),
\label{eq:focal-loss}
\end{equation}
where $\gamma$ is the focusing parameter and $w(y)$ is the class weight. The positive-class weight is computed from the ratio of legitimate to fraudulent transactions in the training set. Model checkpoints are selected according to validation AUC-ROC, and the classification threshold is calibrated on the validation set~\cite{niculescu2005calibration}.

\section{Experiment}
The pilot evaluation demonstrates that THGT-FD can extract informative fraud signals from heterogeneous transaction relations and temporal context. The experiment used 150,000 chronologically ordered IEEE-CIS transactions while preserving the original fraud ratio of 3.499\%. This temporal protocol provides a realistic evaluation because the test transactions occur later than those used for model training.

\begin{table}[t]
  \centering
  \caption{Pilot experimental setting on the IEEE-CIS dataset.}
  \label{tab:settings}
  \small
  \begin{tabular}{lr}
    \toprule
    Experimental setting & Value \\
    \midrule
    Sample size & 150,000 \\
    Fraud ratio & 3.499\% \\
    Training transactions & 105,000 \\
    Validation transactions & 22,500 \\
    Test transactions & 22,500 \\
    Processed transaction features & 412 \\
    Numerical features & 390 \\
    Frequency-encoded categorical features & 22 \\
    Transformer layers & 2 \\
    Attention heads & 4 \\
    Hidden dimension & 64 \\
    Time2Vec dimension & 16 \\
    Epochs & 15 \\
    \bottomrule
  \end{tabular}
\end{table}

Given the substantial class imbalance, we focus primarily on AUC-ROC, Average Precision, Recall@1\%, and Recall@5\%. These metrics characterize both global ranking quality and fraud coverage under constrained review budgets. F1 and Accuracy are reported as complementary threshold-dependent metrics. Average Precision is especially informative for imbalanced classification~\cite{saito2015pr}, while top-ranked recall reflects fraud coverage under constrained investigation capacity~\cite{hand2008criteria}.

\begin{table*}[t]
  \centering
  \caption{Validation and test results for the tabular baseline and THGT-FD.}
  \label{tab:results}
  \small
  \setlength{\tabcolsep}{5pt}
  \begin{tabular}{llrrrrrr}
    \toprule
    Model & Split & AUC-ROC & AP & R@1\% & R@5\% & F1 & Accuracy \\
    \midrule
    HistGradientBoosting & Validation & 0.8652 & 0.4700 & 0.2447 & 0.5155 & 0.2933 & 0.8801 \\
    HistGradientBoosting & Test & \textbf{0.8722} & \textbf{0.4681} & \textbf{0.2679} & \textbf{0.4973} & \textbf{0.2525} & \textbf{0.8595} \\
    THGT-FD & Validation & 0.8506 & 0.4475 & 0.2410 & 0.4795 & 0.1371 & 0.6124 \\
    THGT-FD & Test & 0.8536 & 0.4164 & 0.2414 & 0.4708 & 0.1315 & 0.6127 \\
    \bottomrule
  \end{tabular}
\end{table*}

THGT-FD achieved an AUC-ROC of 0.8506 on the validation set and 0.8536 on the test set. Its Average Precision reached 0.4475 and 0.4164, respectively. The test Average Precision remains substantially above the positive-class prevalence of 3.499\%, showing that THGT-FD effectively concentrates fraudulent transactions near the top of the ranked list. Moreover, the nearly identical validation and test AUC-ROC values indicate stable ranking performance across the temporal split.

The strongest advantage of THGT-FD appears in the high-risk review setting. On the test set, Recall@1\% reached 0.2414, meaning that the highest-risk 1\% of transactions contained 24.14\% of all fraudulent cases. Recall@5\% further increased to 0.4708, indicating that a review queue containing only 5\% of the transactions captured nearly half of the fraud. This concentration is particularly valuable in operational environments where investigation resources are limited and analysts must prioritize a small set of high-risk transactions.

HistGradientBoosting achieved the highest aggregate scores, with a test AUC-ROC of 0.8722 and an Average Precision of 0.4681. Nevertheless, THGT-FD remained highly competitive, with an AUC-ROC difference of only 0.0186. More importantly, the difference in high-risk retrieval was limited to 0.0265 for both Recall@1\% and Recall@5\%. These results show that THGT-FD approaches the strong tabular baseline in the practically important task of identifying transactions that warrant immediate review. Validation-stage fusion of a small, diverse set of classifiers offers a complementary direction for improving imbalanced fraud detection~\cite{han2026validationstage}.

This performance is notable because the two models exploit different sources of predictive structure. HistGradientBoosting is well suited to nonlinear interactions and threshold effects in anonymized numerical variables. THGT-FD complements this feature-centric view by explicitly representing card, product, address, email, device, and temporal information as interacting tokens. The model can therefore evaluate each transaction together with the heterogeneous relational context implied by its repeated attributes.

The competitive test performance suggests that relation tokens provide a compact and effective mechanism for incorporating relational information without constructing a full transaction graph. Relation-specific embeddings preserve the semantics of different entity types, while self-attention models their interactions with the transaction representation. Time2Vec further introduces a learnable temporal signal, allowing the model to represent both gradual changes and recurring activity patterns along the relative transaction timeline.

The threshold-dependent results provide an additional opportunity for deployment-oriented improvement. THGT-FD already produces informative risk rankings, as demonstrated by its AUC-ROC and top-ranked recall. Its binary prediction performance can therefore be further strengthened through probability calibration and validation-based threshold selection. In practice, selecting the threshold according to review capacity may better align the model output with the operational objective than applying a fixed threshold of 0.5. This emphasis on downstream utility is consistent with risk-aware stochastic decision policies and cost-sensitive hierarchical fusion in adjacent financial applications~\cite{song2023deterministic,kou2026learningtofuse}. More broadly, work on tabular foundation models for discrete choice shows that predictive accuracy alone need not guarantee economically valid decisions and that structural constraints can preserve domain-consistent behavior~\cite{wang2026economicvalidity,wang2026embedding}.

\begin{figure}[t]
  \setlength{\abovecaptionskip}{3pt}
  \centering
  \includegraphics[width=\columnwidth]{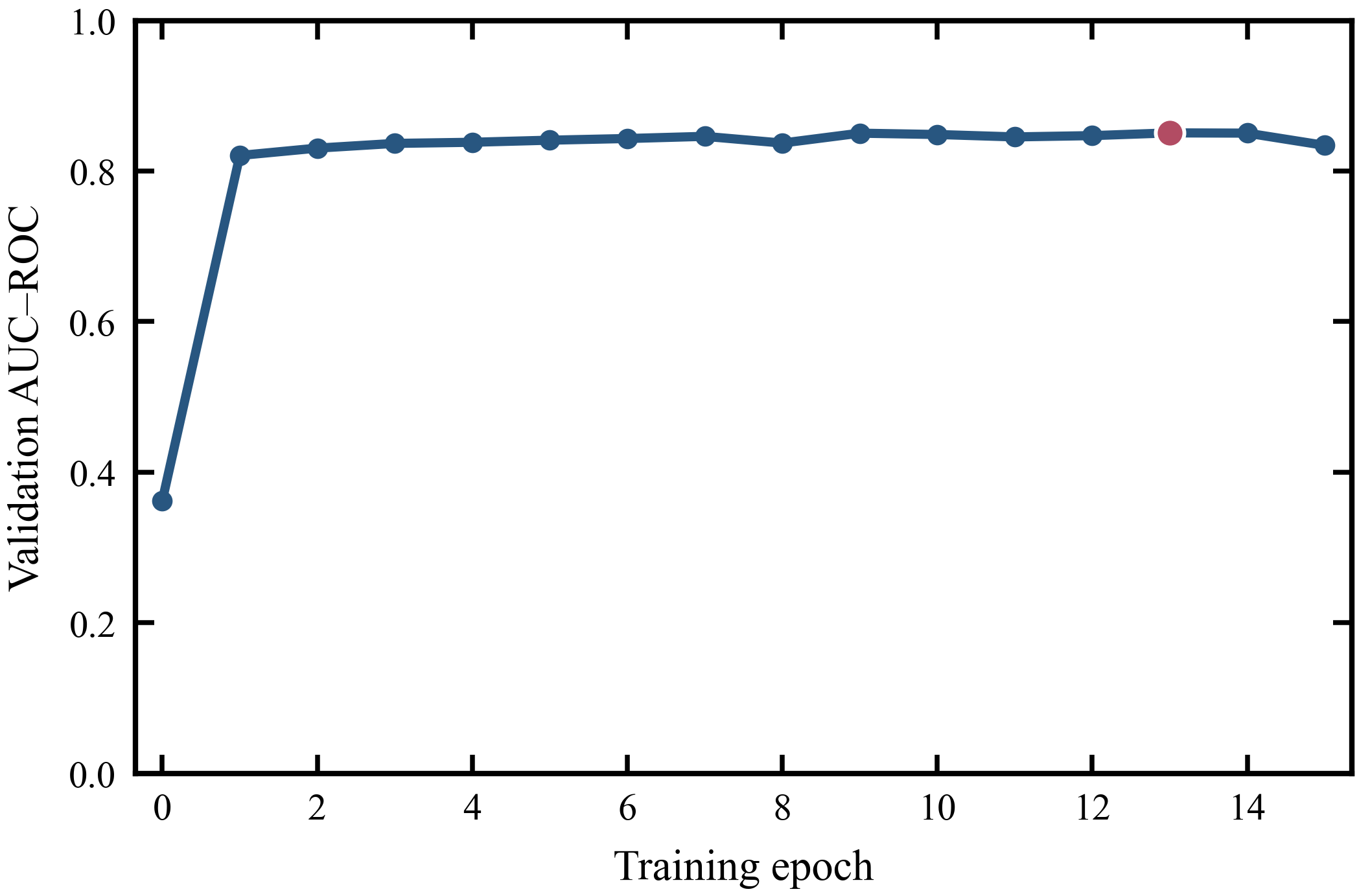}
  \Description{A line chart showing validation AUC-ROC across THGT-FD training epochs.}
  \caption{Validation AUC-ROC across THGT-FD training epochs.}
  \label{fig:training}
  \vspace{-8pt}
\end{figure}

The training curve shows rapid initial learning followed by moderate epoch-to-epoch fluctuations. Validation AUC-ROC increased from 0.3623 before optimization to 0.8206 after the first epoch and reached its maximum of 0.8506 at epoch 13. The subsequent decline supports retaining the epoch-13 checkpoint rather than the final training state.

Focal loss and positive-class weighting contribute to the model's sensitivity to rare fraudulent transactions. By assigning greater importance to difficult and minority-class examples, these components support the strong fraud coverage observed in the highest-risk review groups. Further tuning of the focal parameter and class weights may provide an even better balance between ranking quality, probability calibration, and threshold-based performance; recent dollar-metric evidence also suggests that class weighting, amount-conditioned weighting, and post-training alert reranking can produce different operational trade-offs~\cite{wu2026classweighting}.

A particularly promising aspect of THGT-FD is that its current performance is obtained using a compact transaction-centered architecture. The model does not yet require explicit multi-hop graph propagation or historical aggregation across all transactions connected to an entity. Even under this lightweight formulation, it captures meaningful interactions among cards, devices, addresses, email domains, products, and temporal signals. This provides a scalable foundation for extending relational modeling without abandoning the efficiency of token-based processing.

Entity-level historical aggregation represents a natural next step. Incorporating recent activity associated with a card, device, address, or email domain could enrich each relation token with behavioral context. Such information may strengthen the representation of repeated device use, emerging card-device combinations, and short-term transaction bursts. Tail-sensitive summaries of extreme activity may also be worth testing, given evidence that jump tail risk can improve prediction in the distinct but related task of financial-distress assessment~\cite{liu2023financialdistress}. Multi-hop relational interactions could further propagate risk information across transactions connected through shared entities. Beyond transaction histories, knowledge-enhanced financial forecasting distinguishes graph-based and non-graph-based context and provides useful designs for fusing external knowledge with historical features~\cite{wang2026knowledge}.

The pilot sample may capture only a subset of the recurrence patterns available in the complete IEEE-CIS dataset. Larger-scale training would provide more observations for low-frequency relation values and support more stable entity embeddings. It would also allow the model to learn a broader range of interactions between transaction attributes, temporal behavior, and heterogeneous relations. Under shifts across time periods or deployment populations, sparse causal discovery and graph-domain adaptation could help separate stable signals from domain-specific correlations~\cite{luo2025slogan}. Component-level ablations could then quantify the individual contributions of relation tokens, Time2Vec, and attention-based interaction modeling.

Overall, THGT-FD provides a promising framework for integrating heterogeneous relations and temporal information into transaction fraud detection. It delivers competitive AUC-ROC, strong fraud coverage under restricted review budgets, and stable ranking performance across the temporal split. These findings demonstrate the value of transaction-centered relational modeling and establish a strong basis for richer historical aggregation, calibrated decision-making, and large-scale evaluation.

\section{Conclusion}
This paper introduced THGT-FD, a temporal heterogeneous Transformer for credit card fraud detection on the IEEE-CIS dataset. Each transaction is encoded as a compact sequence that combines a feature token with a Time2Vec representation. Relation tokens describe the shared entities associated with the transaction. This formulation incorporates relational context while retaining efficient token-based processing. In the pilot experiment, THGT-FD achieved a test AUC-ROC of 0.8536 and captured 47.08\% of fraudulent transactions within the highest-risk 5\%. These findings demonstrate the viability of temporal heterogeneous relation modeling for risk-prioritized fraud screening. The model also remained competitive with a strong class-weighted gradient-boosting baseline.

Future work will enrich relation tokens with entity histories and rolling temporal statistics. Multi-hop propagation could further support information exchange between transactions connected through shared entities. Full-data training will provide broader relational coverage, while targeted ablations can quantify each component's contribution. Probability calibration will also align the decision threshold with operational review capacity.

\bibliographystyle{unsrt}
\bibliography{references}

\end{document}